\documentclass[journal]{IEEEtran}
\usepackage[utf8]{inputenc}
\usepackage[T1]{fontenc}

\usepackage{silence}
\usepackage{graphicx}
\usepackage[hidelinks]{hyperref}
\usepackage{subcaption}
\usepackage{xcolor}
\usepackage{xstring}

\title{BenchBot: Evaluating Robotics Research in Photorealistic 3D Simulation and on Real Robots}
\author{Ben Talbot, David Hall, Haoyang Zhang, Suman Raj Bista, Rohan Smith, Feras Dayoub, and Niko S\"{u}nderhauf\thanks{The authors are with Queensland University of Technology (QUT) in Brisbane, Australia. This research was conducted by the Australian Research Council Centre of Excellence for Robotic Vision (project number CE140100016), and supported by the QUT Centre for Robotics. Email: \href{mailto:b.talbot@qut.edu.au}{\nolinkurl{b.talbot@qut.edu.au}}}}
\date{June 2020}

\begin{document}

\maketitle

\begin{abstract}
  We introduce \textit{BenchBot}, a novel software suite for benchmarking the performance of robotics research across both photorealistic 3D simulations and real robot platforms. BenchBot provides a simple interface to the sensorimotor capabilities of a robot when solving robotics research problems; an interface that is consistent regardless of whether the target platform is simulated or a real robot. In this paper we outline the BenchBot system architecture, and explore the parallels between its user-centric design and an ideal research development process devoid of tangential robot engineering challenges. The paper describes the research benefits of using the BenchBot system, including: enhanced capacity to focus solely on research problems, direct quantitative feedback to inform research development, tools for deriving comprehensive performance characteristics, and submission formats which promote sharability and repeatability of research outcomes. BenchBot is publicly available\footnote{Available on GitHub, via \url{http://benchbot.org}}, and we encourage its use in the research community for comprehensively evaluating the simulated and real world performance of novel robotic algorithms.
\end{abstract}
% comprehensively benchmarking the \textit{semantic scene understanding} performance of systems. The performance is evaluated on two tasks---semantic SLAM and scene change detection---both requiring a task solution to develop an object-based semantic map of the environment. A novel evaluation metric called \textit{object map quality} (OMQ) is also introduced for evaluation of these semantic scene understanding tasks with BenchBot.
\begin{IEEEkeywords}
  benchmarking research, research evaluation, 3D simulation, robotics software
\end{IEEEkeywords}

\section{Introduction}

\begin{figure}[p]
  \centering
  \includegraphics[width=0.9\linewidth]{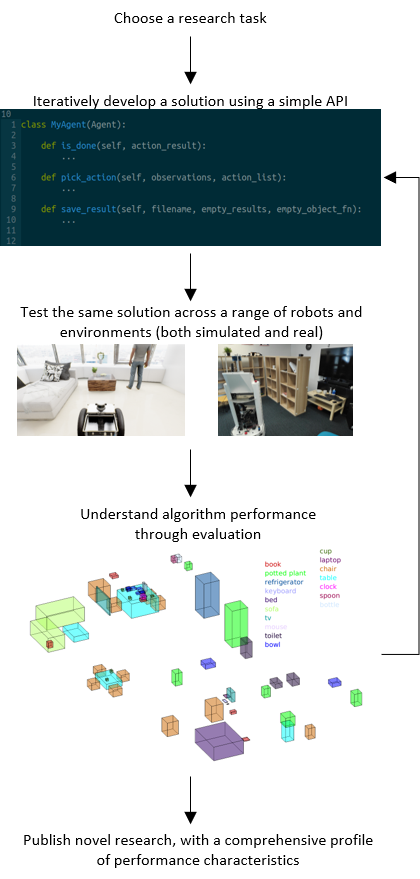}
  \caption{The research process enabled by the BenchBot system}%
  \label{fig:benchbot}
\end{figure}

Robotics research requires comprehensive evaluation prior to being deployed on real systems to guarantee robustness, a characteristic crucial for robots that operate with humans in the real world. Guaranteeing the robustness of robotic systems in the real world is challenging due to the dichotomy between the deterministic computing environment in which algorithms are conceived and the unpredictability of the real world. Consequently, evaluation processes that operate both within simulated computing environments and real world applications play a crucial role in determining the robustness of novel research.

Standardised benchmarks have become synonymous with evaluating performance in fields like computer vision, but have had limited adoption in the robotics field due to challenges associated with standardising robotics~\cite{del2006benchmarks}. Fundamentally, standardising robotics problems diverges from a number of the key characteristic that define robotics research---e.g. robustness under changing environments, lighting conditions, climates, sensors, and platforms.

The research community has taken varying approaches in trying to reliably evaluate the performance of robotics research outcomes. These include static datasets like the Oxford RobotCar dataset~\cite{RobotCarDatasetIJRR} that ignore the agency of a robot, engineering-heavy real world competitions like the DARPA robotics challenges~\cite{krotkov2017darpa}, game-engine powered high fidelity simulations like AirSim~\cite{airsim2017fsr}, and 3D environment reconstructions from real world data~\cite{xia2018gibson}. Although there is a wide range of approaches, a standard approach and toolset for comprehensively evaluating the performance of robotics research has not yet been established.

We present the BenchBot system (shown in Fig. \ref{fig:benchbot}) as a software tool for comprehensively evaluating the performance of novel robotics research in both high-fidelity 3D simulation and on real robots.
The system allows users to define tasks their research is trying to solve, declare metrics for evaluation of performance, and integrate their research with the sensorimotor capabilities provided by robotic systems. The software suite seamlessly transitions between evaluation in simulation and the real world to facilitate comprehensive testing of novel research systems. This paper describes the following key contributions provided by the BenchBot software suite:
\begin{itemize}
  \item a simple Python API for interaction with the underlying robot system,
  \item support for complex changes in target scope (i.e. research task, robot platform, and operating environment),
  \item ability to run the same research on both simulated and real robot platforms without code changes,
  \item a customisable evaluation pipeline for guiding research development through quantitative feedback,
  \item a batch operation mode for building comprehensive performance profiles of novel research algorithms, and
  \item modular design for easy extension to new research tasks, robot platforms, and operating environments.
\end{itemize}

% Insert bit about the semantic scene understanding case study / challenge here:
% \begin{itemize}
%   \item BenchBot, an open source software suite for the comprehensive evaluation of novel robotics research across simulation and real world applications;
%   \item An evaluation framework for semantic scene understanding research, including the definition of two tasks and a challenge to the community\footnote{The Semantic Scene Understanding challenge is available through EvalAI \ben{CITE} at \ben{\url{https:MAKE_A_TINY_URL}}};
%   \item Object map quality (OMQ), a novel metric for evaluating the quality of an object-based semantic map; and
%   \item A demonstration of constructing a solution to the semantic scene understanding tasks, and using the evaluation capabilities provided by the BenchBot software suite to optimise the performance of the solution.
% \end{itemize}

The rest of the paper is organised as follows. Section \ref{sec:related} discusses existing approaches to benchmarking and evaluation in robotics, and methods for robot simulation. Next, the BenchBot system and its underlying components are formally described in Section \ref{sec:system}. The paper concludes in Section \ref{sec:conclusion} with a discussion of the results and future intentions for the BenchBot system.

% A case study is performed with BenchBot on the semantic scene understanding research problem in Section \ref{sec:study}, with the OMQ metric and a basic solution described.

\section{Related Work}
\label{sec:related}

% \ben{The following paper is a good reference point for creating our related work section: \url{https://arxiv.org/pdf/1910.14442.pdf}}.

We provide context for the BenchBot system by outlining the current standards for benchmarking and evaluation within robotics, and then look specifically at robot simulation processes meant to enable benchmarking.

\subsection{Benchmarking and Evaluation in Robotics}
Standardized benchmarks are not common in robotics research when compared to fields like computer vision.
This can largely be attributed to how difficult standardizing a robotics test can be, requiring standardized hardware, software, and environments~\cite{del2006benchmarks}.
This has led to a culture of robotics testing via experimentation to prove a hypothesis rather than comparison and evaluation~\cite{corke2020can}.
Corke et al.~\cite{corke2020can} postulates that this is a factor which limits the speed of progress in robotics research in comparison to similar fields which instead perform regular evaluation and comparison of techniques.
Despite being uncommon, there are some typical approaches used when trying to create benchmarks for robotic systems.

The first approach is to use pre-recorded data and evaluate how well algorithms interpret that data for specific tasks.
Some well-known examples of this are the KITTI~\cite{Geiger2013IJRR}, Cityscape~\cite{Cordts2016Cityscapes} and Oxford RobotCar~\cite{RobotCarDatasetIJRR} datasets which enable tasks such as object tracking, visual odometry, SLAM, and semantic segmentation to be evaluated.
While this approach drives research in data interpretation, it loses the active nature of robotics wherein observations inform actions to solve problems.

Another approach to standardize robotics testing is to provide a consistent environment for robotic testing while leaving other variables of robot design open.
This enables both active interaction with the environment, and different hardware and software solutions to be tested and compared.
% \david{I feel like there are other big ones I might be missing. Please check/give opinions}
This approach is seen by competitions like RoboCup@Home~\cite{rulebook_2019}, the DARPA robotics challenge~\cite{krotkov2017darpa}, and the Amazon picking challenge~\cite{correll2016analysis}.
While enabling system comparison, we see three main limitations to this approach.
Firstly, these events are too infrequent to drive research in the same manner seen in computer vision and machine learning research.
Secondly, while good for systems-level comparison, precise research outputs (algorithms, sensor design, etc.) cannot be easily compared.
% \david{Wanted to put in something about research vs. clever engineering but feels out of place (see commented old version of above sentence).}
% Secondly, while this method allows for systems-level analysis, it can be unclear what specific research outputs have enabled given results (algorithms, new sensors, etc.) vs. well disciplined engineering.
Finally, these competitions become monetarily restrictive with groups needing access to their own physical robotic platforms, large engineering investment, and significant transport funding.
% Finally, they are restrictive to only those who have access both to their own physical platform and engineers capable of making the whole system work, turning the problem into something of a game of money.

An interesting approach being newly adopted, is to provide remote access to robot platforms.
This is present in challenges like RoboThor~\cite{deitke2020robothor}, iGibson~\cite{xia2020interactive} and the Real Robot Challenge\footnote{https://real-robot-challenge.com/en}.
This avenue presents promise in giving users access to physical hardware without the costs of setup and maintenance of the hardware for most users.
While not currently a highly scalable solution, using real robot platforms provides the most realistic real-world performance whilst enabling interaction.
The issue of scalability can be lessened somewhat by combining remote access to real platforms with the use of high-fidelity simulations.

% The final approach we consider here for standardizing robotics research is to use simulation to approximate real-world environments, conditions and sensors.
% Here, the robotics platform and environment can be standardized whilst still enabling active robotic interaction.
% The techniques applied for this problem are explained in the following section.
% \ben{This is important, but I don't have a clear picture in my head of what points need to be made. The general point is to finish with a gap clearly established: benchmarking and evaluation systems in robotics are typically limited to static datasets that don't cover the full breadth of the experiences a robot system needs to handle (i.e. simulation for covering a wide range of experiences, real world for actually working in the real world, etc.)\ldots}

% \david{Plan to take inspiration from Peter's paper on this. Essentially stating the success of computer vision systems in comparing and benchmarking whereas robotics has a lack in this area. Moving from experimentation to evaluation. Examples given of evaluation could be KITTI and several small SLAM datasets (need to find some good ones). Some good references in the iGibson paper. Remember to add active vision dataset to this discussion}

\subsection{Robot Simulation}
Robot simulation endeavours to provide tools that enable consistent robotics testing.
We define two approaches to this which are utilised within the literature.

The first approach is the creation of fully simulated environments where the environments are hand-crafted and designed, typically through using a game engine.
Some well known examples of this are AirSim~\cite{airsim2017fsr} and CARLA~\cite{Dosovitskiy17} for outdoor environments, and AI2Thor~\cite{kolve2017ai2}, RoboThor~\cite{deitke2020robothor} and Isaac\footnote{\url{https://www.nvidia.com/en-us/deep-learning-ai/industries/robotics/}} for indoor environments.

The second approach is the creation of simulated environments by utilizing real-world data.
Generally this comes in the form of a full 3D environment that a simulated agent can explore freely, such as is seen by AIHabitat~\cite{habitat19iccv}, Gibson~\cite{xia2018gibson}, and iGibson~\cite{xia2020interactive}.
Functionally these can act identically to those created in a games engine but generated using sets of depth and image data collected throughout the environment, which are stitched together to create the final simulated environment.
Alternatively, real-world data can be used to provide precise real-world sensor readings from specific pre-defined poses within an environment.
% \david{Should double check how this data is meant to be used (simulator attached etc.)}
This is the approach of the active vision dataset~\cite{active-vision-dataset2017}, which, while not a simulator in the same sense of the others shown here, enables ``traversal'' between densely sampled poses to simulate movement while providing the precise visual data captured at that location.
% An alternative approach, while not technically a full simulation, is the approach of the active vision dataset~\cite{active-vision-dataset2017} where data is collected at predefined nodes of set increments apart and an agent can move between them to simulate movement, whilst always receiving the precise data that was captured at that point.

\begin{figure}[t]
  \centering
  \begin{subfigure}[b]{0.48\linewidth}
    \includegraphics[width=\textwidth]{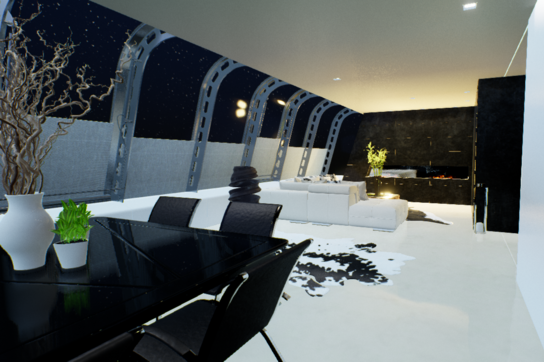}
    \caption{}
  \end{subfigure}
  \begin{subfigure}[b]{0.48\linewidth}
    \includegraphics[width=\textwidth]{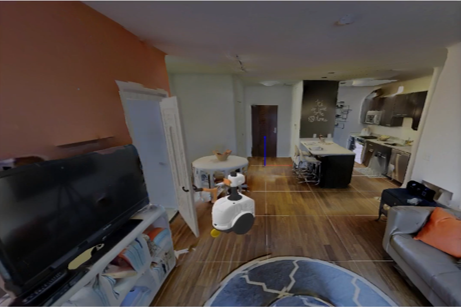}
    \caption{}
  \end{subfigure}

  \caption{Example output from two different approaches to robot simulation: a) the probabilistic object detection challenge~\cite{skinner2019probabilistic} (is fully simulated with realistic light reflections, but for a spaced out ``clean environment''), and b) iGibson~\cite{xia2020interactive} (built from real-world data real-world data with realistic clutter and textures, but contains some visual artefacts).}
  \label{fig:sim_compare}
\end{figure}

There are distinct advantages and disadvantages when using either fully simulated or real-world data simulators.
On a practical level, fully simulated environments are easily manipulated and adapted for new conditions (e.g. lighting variations, rearranging objects, etc.).
This is more challenging for simulated environments created from manually observing real-world environments.
However, manually collecting data typically provides more naturally messy environments with randomly cluttered surfaces.
This seems more realistic when compared to the clean and spacious environments given by many fully simulated environments.
% For example, the natural mess of environments captured by using real-world data is typically not seen in simulated datasets, with spaces looking very clean and spaceous without random clutter on surfaces.
% However, in manually collecting such data, adapting the environments for new conditions (e.g. lighting variations, rearranging objects, etc.) becomes extremely challenging.
% This is not a large issue for fully simulated simulators.
% However, this restricts the freedom in adapting the environment for new conditions which is easier done in fully simulated environments (e.g. lighting variations, rearranging objects, etc.) without needing to manually collect and label all data from the same environment multiple times.
% \david{Rephrase this half later.}
However, the subject of visual realism between approaches is a hotly contested topic.
Without fixed agent poses such as those used in the active vision dataset~\cite{active-vision-dataset2017}, using real data to create a virtual environment can lead to visual artefacts being introduced and realistic lighting reflections cannot be achieved as lighting is not actively calculated.
While this is not an issue for fully simulated environments, current robotics simulators of this type don't typically have particularly realistic textures (unlike when using real-world data) which can leave object appearances looking ``flat''.
You can see a simplistic comparison of simulator approaches in Figure~\ref{fig:sim_compare}.
Regardless of the approach used, it is important to accept that a sim-to-real gap will always be present when using simulators.

The sim-to-real gap is perhaps best addressed when combined with real-world robot platforms.
This is the approach used by RoboThor~\cite{deitke2020robothor} and iGibson~\cite{xia2020interactive} that can directly examine the performance degradation in sim-to-real transfer.
This enables rapid repeatable prototyping in high-fidelity simulation while still enabling direct analysis of real world performance.
This is also a key component used in the design of our BenchBot system.

% \suman{May be good to have comparison table of different platform from David's Presentation.}

% \ben{The paper mentioned above has a good section on this. We need to describe / classify the different methods of simulation, talking about the advantages and disadvantages of each. A key distinction is game engine renderers vs mesh renderers (i.e. ``using 3D capturing methods to scan real environments'').}
% \david{Systems to discuss:
% \begin{itemize}
%     \item AI2Thor/RoboThor
%     \item Gibson/iGibson
%     \item AIHabitat
%     \item Isaac
%     \item AirSim and Carla?
%     \item Active Vision Dataset?
% \end{itemize}
% Main points to bring up are visual fidelity, abstract APIs for non-roboticists, and sim-to-real transfer.}

\section{The BenchBot System}
\label{sec:system}

Robotics researchers using the BenchBot system are able to focus on the research process---defining problems, creating solutions, and improving results---without being encumbered by the complications that underpin complex robotic systems. The BenchBot system is a software suite that manages the process of applying entire robot systems to a variety of research tasks, regardless of if the robot systems are simulated or operate in the real world. The software suite provides this wide scope of capabilities, while prioritising minimising the configuration and interaction burdens passed to the end-user.

\begin{figure*}[t]
  \centering
  \includegraphics[width=0.9\linewidth]{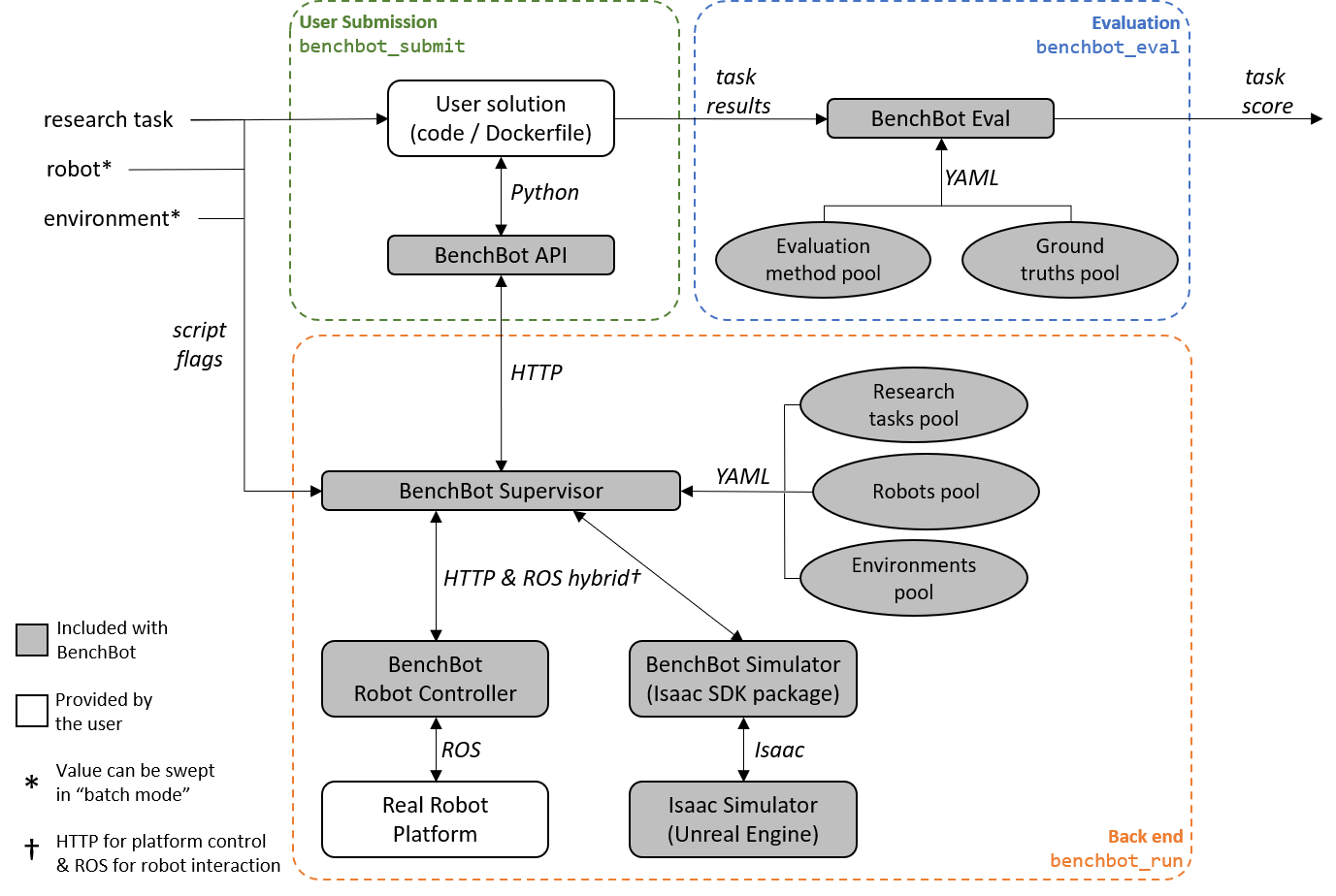}
  \caption{Architecture of the BenchBot system. A user creates a submission that attempts to perform a defined research task, and receives an evaluation of their performance in return. Users write code that interfaces with a simple API, and the back end internally handles interaction with the underlying simulated or real robot platform.}%
  \label{fig:system}
\end{figure*}

BenchBot provides three scripts for using the system, which are denoted by the coloured sections of the system architecture shown in Fig. \ref{fig:system}. The scripts allow users to: 1) select a research task, robot platform, and environment to run; 2) submit a solution for the research task; and 3) obtain evaluation feedback  to iteratively improve their solution's performance. Each of these three steps directly map to the steps of the ideal research process described above: 
defining problems, creating solutions, and improving results.

The user simply employs these scripts in their research process, which BenchBot then uses to manage all of the complex underpinnings required of a robot system. Each area of the BenchBot system managed by these scripts is discussed in further detail in Sections \ref{subsec:backend}, \ref{subsec:api}, and \ref{subsec:eval} respectively, along with their underlying components. Lastly, this section concludes in Section \ref{subsec:batch} with a discussion of batching---BenchBot's tool for effortlessly building comprehensive performance profiles of robotics research.

\subsection{Declaring a problem with the BenchBot back end}
\label{subsec:backend}

Using BenchBot begins by clearly declaring a problem that needs to be solved, a step synonymous with beginning research. The back end of the BenchBot system (including entire robot platforms, simulators, configuration, initialisation, networking, and interfacing) is started through a single script called \texttt{benchbot\_run}. The script requires the user to select a target research task, robot platform, and operating environment from the pool of available options. Supported options are listed through helper flags, and the script validates whether the selected configuration is achievable (e.g. running a real robot in a simulated environment is not a valid configuration).

Options are declared to the BenchBot back end simply by creating a YAML file in the appropriate pool describing the configuration option, and providing any necessary data described by the configuration. For example, a simulated environment definition would contain the data for the environment simulation and a YAML file declaring an identifier, the type of environment (simulated or real), a start pose for the robot, a trajectory the robot may use to travel through the environment, etc. Robots are declared as a series of directed connections between robot platform and BenchBot API along with functions for translating data between the two endpoints, and tasks are declared as a list of available robot capabilities.

\subsubsection{BenchBot supervisor} is started once a valid selection has been provided. The supervisor serves as the central component of the BenchBot back end, providing a single interface to handle the conglomeration of data required to manage the robot system: command line selections, HTTP communication with the BenchBot API, configuration YAML files, ROS sensorimotor data, environment initialisation commands, and HTTP control commands for simulators and real robots. In handling this wide range of data, the supervisor is able to load data for the selected configuration from the available pools, manage the life cycle of the underlying robot platform and environment (whether that be real or simulated), and provide a conduit between the sensorimotor capabilities of the robot and the simplicity of access provided by the BenchBot API.

\subsubsection{Real robot platforms} sit below the supervisor in the back end architecture, and follow a pattern similar to typical ROS systems. BenchBot adds a robot controller to facilitate the ease-of-use functionality provided by the BenchBot API. Examples include stopping collisions before commands provided by the API can create them, guiding the robot through static trajectories for easier tasks, and returning the robot to consistent starting pose between trials.

\begin{figure}[p]
  \begin{subfigure}{\linewidth}
    \centering
    \includegraphics[width=0.5\linewidth]{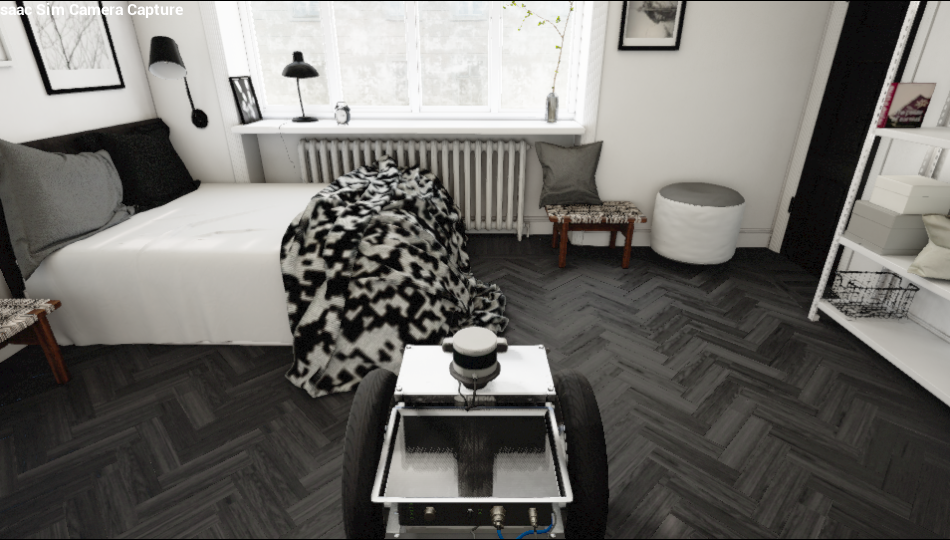}%
    \hfill
    \includegraphics[width=0.5\linewidth]{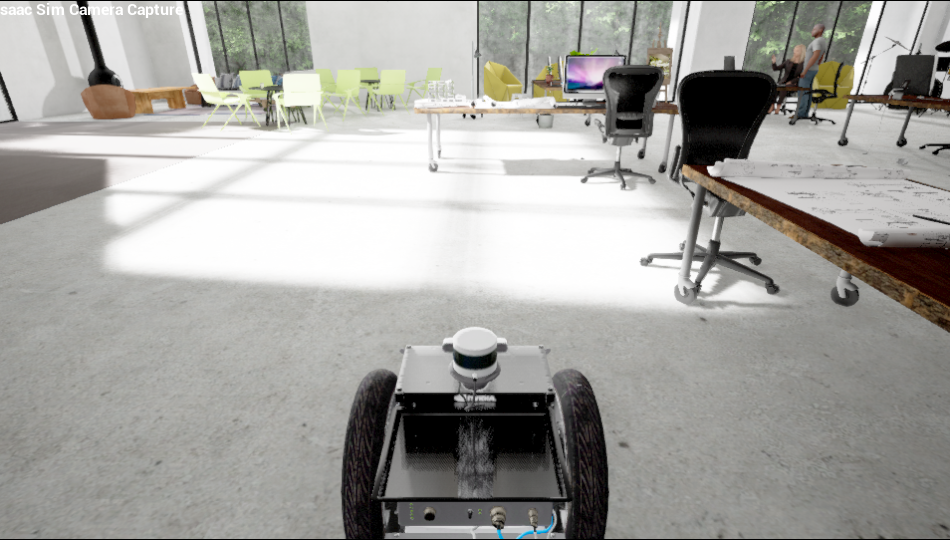}
    \caption{}
    \label{subfig:sim_scale}
  \end{subfigure}
  \begin{subfigure}{\linewidth}
    \centering
    \includegraphics[width=0.5\linewidth]{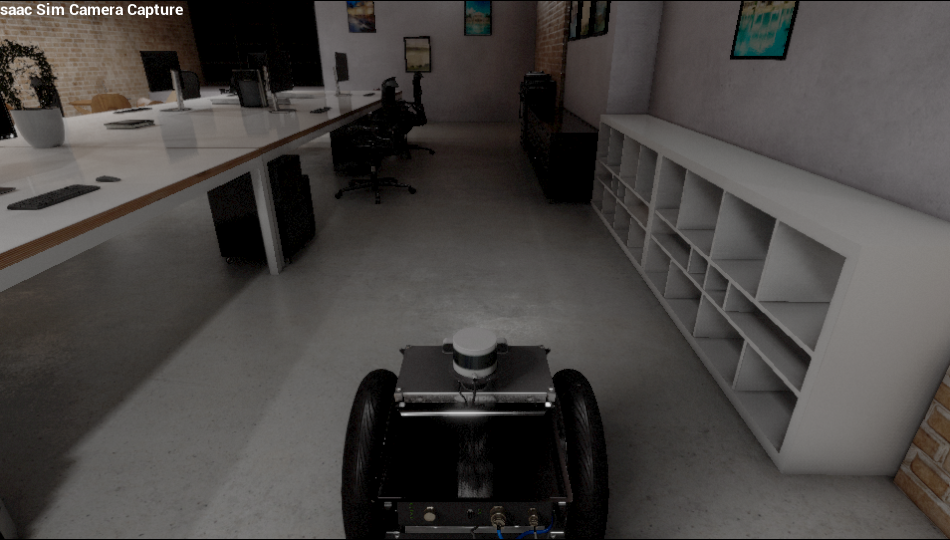}%
    \hfill
    \includegraphics[width=0.5\linewidth]{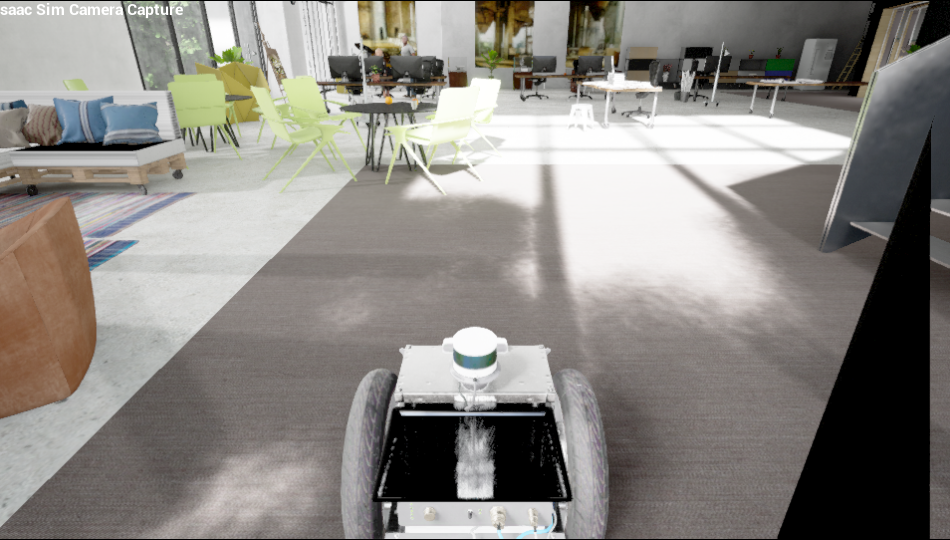}
    \caption{}
    \label{subfig:sim_lighting}
  \end{subfigure}
  \begin{subfigure}{\linewidth}
    \centering
    \includegraphics[width=0.5\linewidth]{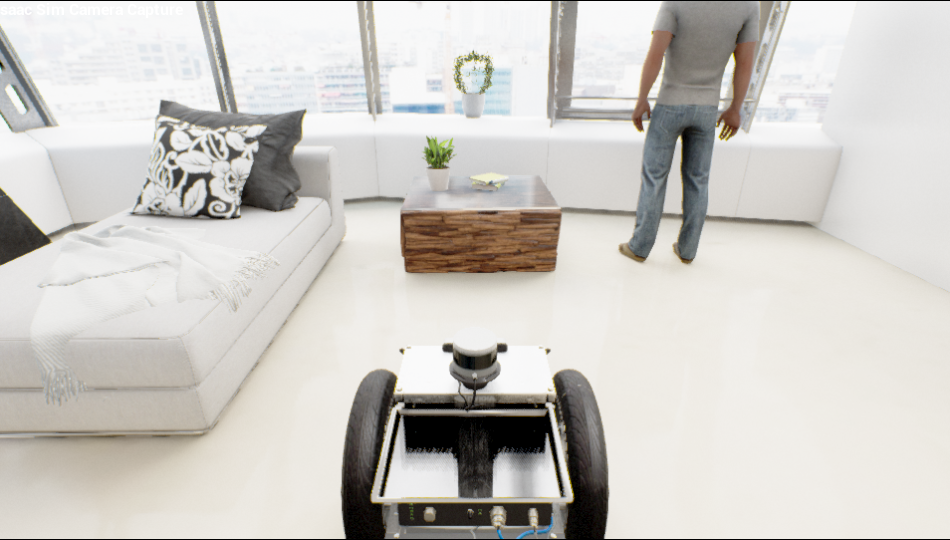}%
    \hfill
    \includegraphics[width=0.5\linewidth]{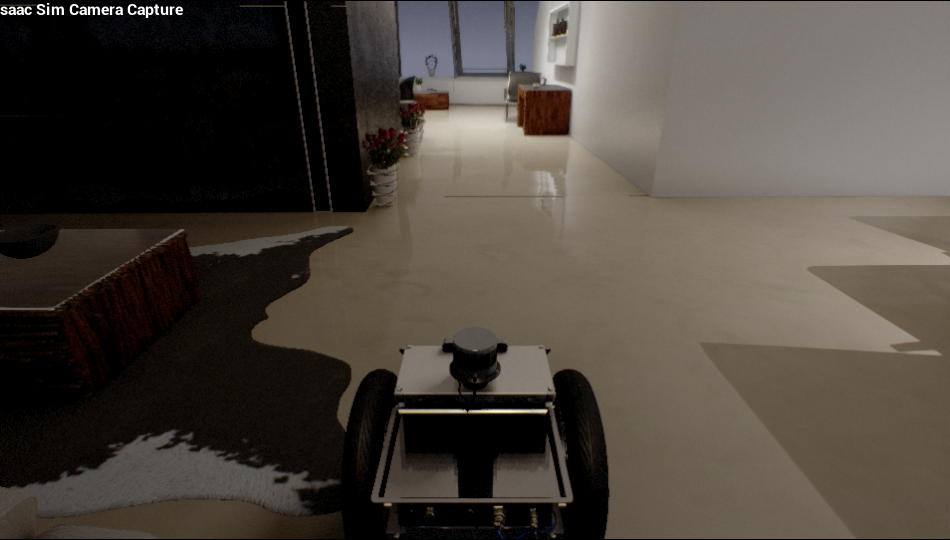}
    \caption{}
    \label{subfig:sim_time}
  \end{subfigure}

  \caption{A sample of the high-fidelity environments generated in Unreal Engine with Nvidia Isaac, with the following variations highlighted: \subref{subfig:sim_scale}) scale, \subref{subfig:sim_lighting}) lighting, and \subref{subfig:sim_time}) time of day.}
  \label{fig:sim}
\end{figure}

\subsubsection{Simulated platforms} in the BenchBot back end heavily leverage the capabilities provided by the NVIDIA Isaac SDK and Isaac Unreal Engine Simulation platform~\cite{Isaac}. The Isaac simulator provides agency of a robot platform within an environment simulated by Unreal Engine's powerful simulation capabilities. The simulator capabilities allow the robot to be simulated in a wide variety of environments that can range in scale, lighting conditions, and even time of day (as shown in Fig. \ref{fig:sim}). A BenchBot simulator package is used above the Isaac components to transform the Isaac robot interface into ROS, and control aspects of the simulator life cycle like restarting from a clean state, declaring robot collisions, and dynamically changing environment selections.

\subsection{Creating research task solutions with the BenchBot API}
\label{subsec:api}

The next step in the research process is creating a solution to the research problem, which once again has an analogous step in the BenchBot process. The user creates a solution that uses the BenchBot Python API to interact with sensorimotor robot capabilities, obtain back end configuration details, and generate structured task results. A user solution is submitted to a running BenchBot back end using the \texttt{benchbot\_submit} script, which supports both native (i.e. running a Python script locally) and containerised (i.e. building a Docker image from a Dockerfile) submission modes. Containerised research solutions can run independently on other systems, enabling easy access and verification of solutions by the research community. Shareable and repeatable research outcomes, a key contribution of the BenchBot system, are a significant driver of progress in the research community.

Design of the BenchBot API is inspired by the ``observe and act'' framing employed in areas of robotics like reinforcement learning and the OpenAI Gym ecosystem~\cite{OpenAIGym}. The API uses data in the task definition to provide a list of sensor observations and possible robot actions to the user. A solution can either combine these manually into a control loop, or declare an agent with the three capabilities required to complete a BenchBot task: choosing an action given a set of observations, knowing when the task is done, and saving results for the task. Breaking the entire process of solving a complex robotics task down into simply providing three functions is an embodiment of the directness in which research can be conducted with the BenchBot system.

\subsection{Measuring performance with BenchBot evaluation tools}
\label{subsec:eval}

Once results have been attained through the BenchBot system, the final step is to evaluate the performance of the solution given the generated results. A \texttt{benchbot\_eval} script is provided to pass a collection of results to the underlying Python evaluation module. The evaluation module supports scoring results individually and producing a summary score for multiple results.

An appropriate evaluation method is selected from the pool of available methods based on the task identifier provided with the results. This flexibility recognises that different tasks will have different metrics that best capture their performance, while also consolidating metrics into single reusable implementations rather than each researcher creating their own implementation.

\subsection{Building comprehensive performance profiles with batches}
\label{subsec:batch}

Although the BenchBot system makes generating a result simple, we recognise that a single result is rarely enough to gain a comprehensive understanding of an algorithm's performance. BenchBot provides a final script \texttt{benchbot\_batch} that generates a set of task results by sweeping over a set of different environment and robot combinations. The script, in combination with evaluation tool support for multiple results, allows a user to produce a comprehensive performance profile for their research contribution with a single command. A performance profile comprising of results from multiple varying simulated environments, multiple robot platforms, and real world results empowers researchers to glean more comprehensive and meaningful insights from their research.

\section{Conclusions \& Future Work}
\label{sec:conclusion}

In summary, we have described the BenchBot system from an architectural level and explored the capabilities the user-centric system design affords researchers. BenchBot provides simple tools for: targeting a multitude of research tasks, robot platforms, and environments; interacting with the sensorimotor capabilities of a robot platform; developing and iteratively improving research solutions with quantitative feedback; autonomously generating comprehensive performance characteristics for robotics research; and sharing research solutions to promote repeatability and accessibility.

BenchBot is in its infancy, with it currently targeting semantic scene understanding tasks on a limited number of robot platforms. As discussed throughout the paper, the modular system architecture employed in BenchBot allows a wide variety of expansions and improvements to the system. Depending on collaborative interest and development drivers, possible future outcomes we could explore with the BenchBot system include:
\begin{itemize}
  \item using the semantic scene understanding tasks internally to produce novel outcomes in the semantic SLAM and scene understanding research fields;
  \item widening the range of supported robot platforms, particularly those with different actuation capabilities like robot manipulators;
  \item adding support for simulation via NVIDIA Omniverse, a new ray-tracing enabled high-fidelity 3D simulation platform;
  \item exposing resource pools (i.e. research tasks, robot platforms, environments, and evaluation methods) to end-users so they can easily create their own content and expansions; and
  \item providing novel research challenges to the community using the BenchBot platform to stimulate and drive innovation.
\end{itemize}

BenchBot allows researchers to focus on developing novel robotics algorithms without the tangential engineering challenges posed by complex robotic systems. The tools provided with BenchBot facilitate feedback-guided research development, and present users with deep insights into the performance characteristics of their research. We encourage researchers to try BenchBot in their research process, get in contact with us if they have any feedback, and help us enable the development of robust robotics research through comprehensive evaluation.

\bibliography{refs}
\bibliographystyle{ieee}
\end{document}